\documentclass{anstrans}
\usepackage{microtype}
\usepackage[utf8]{inputenc}
\usepackage{amsfonts}
\usepackage{amssymb}
\usepackage{graphicx} 
\usepackage{makecell}

\usepackage[per-mode=symbol,round-precision=3,scientific-notation=false,range-phrase = \text{--}]{siunitx}
\usepackage[hyperindex,breaklinks]{hyperref}
\usepackage{tikz}
\usetikzlibrary{trees}
\usepackage{floatrow}
\usepackage{xspace}

\newcommand*{\ie}{i.e.,\@\xspace}
\newcommand*{\etal}{et al.\@\xspace}

\makeatletter
\newcommand*{\etc}{%
    \@ifnextchar{.}%
        {etc}%
        {etc.\@\xspace}%
}
\makeatother

\usepackage[nameinlink,capitalize]{cleveref}
\DeclareUnicodeCharacter{025B}{\ensuremath{\varepsilon}}

\usepackage[frozencache,cachedir=.]{minted}

\Crefname{lstlisting}{Listing}{Listings}
\usepackage{listings}
\lstdefinestyle{gptStyle}{
    basicstyle=\tiny\ttfamily,
    frame=single,
    breaklines=true,
    moredelim=[is][\color{blue}]{[}{]},
    morecomment=[l]{[Output]},
    commentstyle=\color{red},
    xleftmargin=0em,
    breakindent=0pt
}
\title{Integrating LLMs for Explainable Fault Diagnosis in Complex Systems}
\author{Akshay J.\@\xspace Dave$^{*}$, Tat Nghia Nguyen$^{*}$, Richard B.\@\xspace Vilim$^{*}$}

\institute{
$^{*}$Nuclear Science and Engineering, Argonne National Laboratory, Lemont, IL, ajd@anl.gov
}

\begin{document}
\maketitle
\section{Introduction}

The operation of complex systems has necessitated diagnostic capabilities, particularly as part of a broader autonomous operation framework.
The need for diagnostics is driven by the need to achieve cost efficiency and enhance operators' response. 
However, a diagnostic tool's effectiveness significantly depends on its explainability.
In environments like nuclear power plants, where operators must make informed decisions, the ability to understand and trust the diagnostic information presented is of paramount importance: it is not sufficient to be told that something is wrong; it is crucial to understand \textit{why} and \textit{how} it is wrong, especially to make the most effective corrective actions.

To address this problem, there are two requirements: first, the diagnostics capability itself, and second, a computational agent that can provide explanations about diagnoses.
This work presents a system incorporating both aspects to obtain explainable fault diagnostics.
For the former, we adopt PRO-AID, a model-based diagnostics tool utilizing plant data and physics-based models of the system components.
For the latter, we embed a Large Language Model (LLM).

\subsection{Explainability}

The concept of explainability lacks a universally agreed-upon definition.
While some research suggests that explainability can be achieved through visualizing information flow--such as between layers of deep neural networks \cite{choo2018visual}--others advocate for using heuristics like Pearson correlation rankings to relate inputs and outputs \cite{hong2020remaining}.
These methods may be the only viable options in the context of purely data-driven models.
However, as will be discussed further in \cref{sec:proaid}, our work employs a physics-based diagnostic model, which facilitates deriving the causal relations between potential faults and fault symptoms, allowing us to offer precise rationales for our diagnoses.
The work of Langley \etal \cite{Langley_2017} defined explainability as the ability of an intelligent agent to explain their decisions and the underlying reasoning that led to their choices.
Thus, we define explainability as the capability of a system to enable operators to pose arbitrary questions about diagnoses and receive answers grounded in physical models, sensor data, and logical reasoning.

\subsection{Objectives}

We address three major objectives in this paper.
First, we embed an LLM agent inside a system to explain fault diagnoses to the operator.
A problem with LLMs is that their knowledge is limited to the data they have been trained on, and are susceptible to ``hallucinating'' inaccuracies and presenting them as facts.
Thus, secondly, we address the hallucination issues by constraining the output of the LLM.
Finally, we demonstrate the diagnostics system's capabilities in diagnosing a faulty sensor in an operating molten sodium facility.

\section{Methods}

\subsection{Diagnostics Tool}\label{sec:proaid}

A \textit{consistency-based} framework of model-based diagnosis using physics-based models has been developed and implemented in PRO-AID \cite{NGUYEN2020107767}.
The framework relies on normal (fault-free) models of a system. 
Anomalies due to faults can be detected from the inconsistencies between the normal behaviors predicted by the models and observed data.
Under normal conditions, when the system is fault-free, the observed data must satisfy relations imposed by the fault-free system model.
Such constraint relations between observations and the expected normal behaviors are formally defined as analytical redundancy relations (ARRs) \cite{DEKLEER200325}.
Each analytical redundancy relation may involve a subset of sensor data and certain part of the system.
In PRO-AID, each ARR is represented by an equation establishing the constraint among the involved sensor readings. 
The difference between the two sides of the ARR equation is defined as the \label{word:residual}\underline{residual}.
A non-zero residual would indicate a violation to the ARR, implying at least one of the involved sensors or part of the system is faulted \cite{NGUYEN2020107767}.

One of the diagnostic framework's challenges is constructing a physics-based model of the target system.
At the component level, approximations of the physics, including the mass, momentum, and energy conservation equations, are utilized to formulate the component models. 
Each component model may contain several model parameters to be determined by fitting against training data. 
The model training process for each model has a minimum sensor set requirement.
To allow the construction of diagnostic models for maximum coverage, the concept of \label{word:vs}\underline{virtual sensors} were introduced to be used in place of missing sensors for model training \cite{NGUYEN2022109002}.
Virtual sensors are obtained analytically for unmeasured process variables by solving the conservation laws and related constitutive equations available at the system level.
The validity of each virtual sensor depends on the health status of the components and sensors involved in its underlying analytical solution.

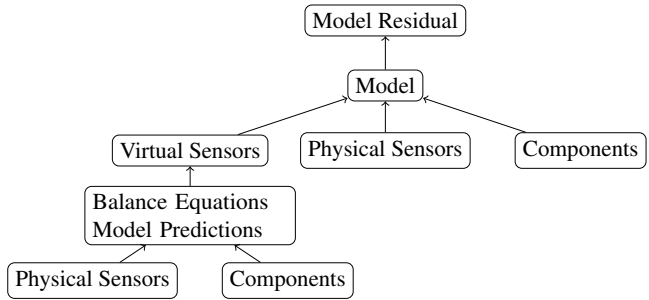
\begin{figure}[h]
  \centering
  \resizebox{\textwidth}{!}{\begin{tikzpicture}[
  grow'=down,
  every node/.style={draw, rectangle, rounded corners=3pt},
  level 1/.style={sibling distance=10mm, level distance=10mm},
  level 2/.style={sibling distance=30mm, level distance=10mm},
  level 3/.style={sibling distance=10mm, level distance=10mm},
  level 4/.style={sibling distance=30mm, level distance=10mm},
  level 5/.style={sibling distance=10mm, level distance=10mm},
  edge from parent/.style={draw, <-, sloped, midway, anchor=center}
]
\node {Model \hyperref[word:residual]{Residual}}
  child {
    node {Model}
    child {
      node {Components}
    }
    child {
      node {Physical Sensors}
    }
    child {
      node {\hyperref[word:vs]{Virtual Sensors}}
      child {
        node [text width=3cm]{Balance Equations Model Predictions}
        child {
          node {Components}
        }
        child {
          node {Physical Sensors}
        }
      }
    }
  };
\end{tikzpicture}}
  \caption{Dependency structure of a generic model residual in PRO-AID.}
  \label{fig:res-structure}
\end{figure}

The set of model residuals forms the basis for fault diagnostics.
Each non-zero residual serves as a fault symptom that implicates certain system faults.
That set of relevant faults, the possible causes of the non-zero residuals, can be identified from the underlying ARR.
Based on the set of all the residuals generated for a system, various reasoning approaches, including both deterministic and probabilistic reasoning, can be employed to produce fault diagnoses\cite{NGUYEN2020107767,NGUYEN2022109002}.
Faults may be detected and diagnosed at a given time based on observed symptoms.
In the presence of measurement and modeling uncertainty, a statistical tool may be needed to determine whether a residual is (statistically) non-zero.
The flowchart in \cref{fig:res-structure} summarizes the dependency structure of a generic model residual.

The \textit{explainability} of the PRO-AID diagnostic results stemmed from the use of physics-based analytical models.
The causal relations between potential faults in the system and possible fault symptoms (non-zero residuals) are derived from the physics-based ARR and stored within PRO-AID.
Fault diagnoses are obtained by logical inference based on observed symptoms, knowing the possible causes of each symptom.
In the reverse direction, the fault diagnosis can be presented along with the observed symptoms and explained intuitively by causality.
Forward chaining can be used by an operator to test that a diagnosis given by the algorithm is logically consistent with the symptoms that led to the diagnosis.
The natural tendency of an operator is to do just that \cite{osti23035374} and can be facilitated by making available the information an operator will use to check for logical consistency. 

To explain the PRO-AID diagnosis, the LLM needs the background information of the physical system and real-time updates of the observations and diagnostic results in PRO-AID.
More specifically, the background information consists of an inventory of physical and virtual sensors, all possible sensor and component faults, and residuals generated for the system.
For each residual, details on its dependency structure, illustrated by \cref{fig:res-structure}, are also provided.
The real-time updates from PRO-AID include recent sensor data, an updated list of zero and non-zero residuals (fault symptoms), and updated diagnostic results.
The data exchange between PRO-AID and the LLM can be done periodically using JSON data files produced by a PRO-AID output module.

\subsection{LLM}

An LLM is a machine learning model that can perform various natural language processing tasks.
LLMs have been embedded in various online portals as chatbots to accommodate arbitrary text-based interactions with humans.
While any LLM could be embedded in the diagnostics framework, we embed GPT-4 developed by \href{https://openai.com}{OpenAI} in this work.
To interact with an LLM, a user must provide strings that have been tokenized -- which is the process of splitting the text into smaller units.
Given that an LLM is pretrained, any external knowledge--about the plant or diagnostics framework--must be provided to it in the form of \textit{context}.
For GPT-4, the context size is 8192 ``tokens''. 
Since a token typically represents approximately three-quarters of a word, this equates to around 12 single-spaced pages of text.
Consequently, carefully managing an LLM's context is imperative to align it with the requisite knowledge for the specific task.

\subsection{Diagnostics System}

\begin{figure}[!hb]
  \centering
  \includegraphics[width=0.9\textwidth]{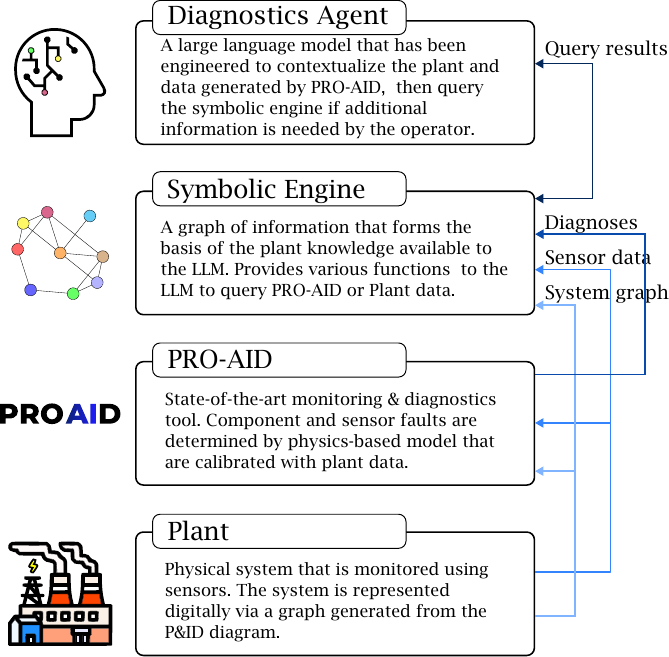}
  \caption{A layout of the system and its interconnections.}
  \label{fig:system}
\end{figure}

To address the objective of integrating LLMs with PRO-AID, we designed a system consisting of four major components: the Plant, PRO-AID, a Symbolic Engine, and the Diagnostics Agent.
The system layout, and the components' description, is presented in \cref{fig:system}. 
The system is designed to create a Symbolic Engine that interfaces between the Agent and the data coming from PRO-AID and the Plant.

The Symbolic Engine is a Python-based package.
Its main function is to manage the context window for the Agent by organizing information about the plant, and analyzing \& presenting sensor data. 
There are two major components of the package.
\texttt{DependencyGraph}: A class to read plant information and then instantiate the underlying knowledge graph of the system.
This graph will store the connections between residuals, faults, components, and sensors.
It will also provide statistical and spectral information about sensor measurements by continuously updating a sensor data buffer.
\texttt{LLMAPI}: A class to embed the OpenAI LLM API and manage the context available to answer queries from the user.
Three functions are available to the user to query information about diagnoses, if any, received from PRO-AID.
These functions will be introduced and demonstrated in \cref{sec:results}.

\subsection{METL Experimental Facility}

\begin{figure}[h]
  \centering
  \includegraphics[width=0.9\textwidth]{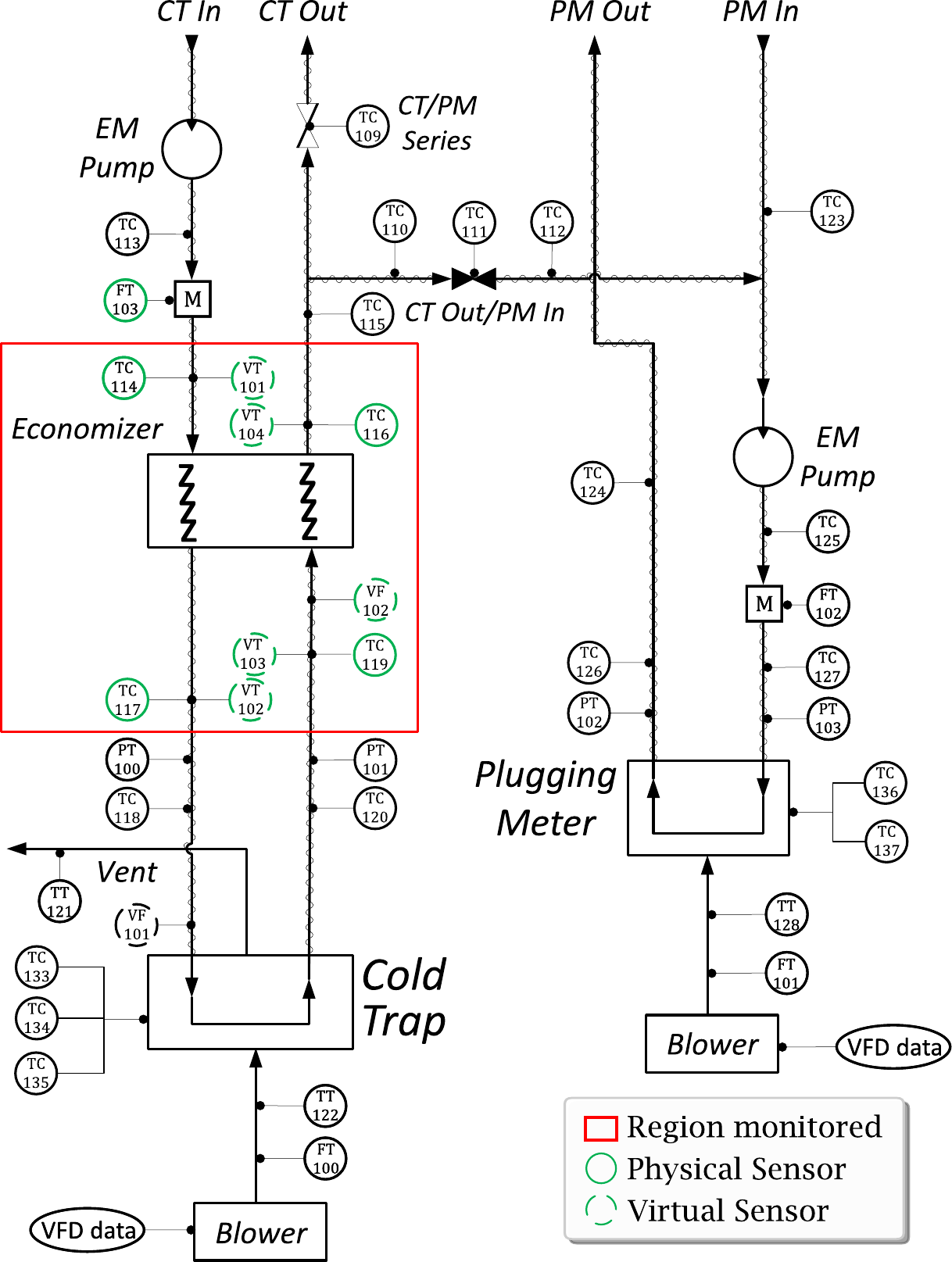}
  \caption{The purification system in METL.}
  \label{fig:METL}
\end{figure}

For a demonstration, we consider applying the proposed framework to the purification loop of the Mechanisms Engineering Test Loop (METL) liquid sodium facility at Argonne National Laboratory.
The sodium purification system, shown in \cref{fig:METL}, consists of two electromagnetic (EM) pumps, an economizer, a cold trap, and a plugging meter.
Each EM pump is equipped with a flow meter. 
The cold trap and the plugging meter are each equipped with a pair of pressure transducers at their inlet and outlet \cite{osti_1492054}. 
Additionally, each piping segment of the system is equipped with a heater and a thermocouple (two thermocouples for piping segments longer than 3 ft) mounted on the outer surface, under an insulation layer, for temperature control.
During operation, the EM pump directs a small fraction of sodium from the main piping system to the cold trap, where the sodium flow is cooled to just above the freezing point.
Due to the reduction of the solubility at lower temperatures, the impurities precipitate out of the sodium flow and adsorb at the stainless-steel mesh filter within the jacket of the cold trap.
An economizer acts as a counterflow heat exchanger to pre-cool the incoming sodium flow to the cold trap and pre-heat the purified outgoing sodium flow.

To monitor the performance of the economizer, since direct measurements of the sodium temperatures are not available, heat balance and heat transfer performance models of the economizer utilizing the thermocouples mounted on the outside piping surface at the inlet and outlet ports of the economizer have been developed and implemented in PRO-AID.
These models and the associated virtual sensors allow PRO-AID to detect and differentiate between component faults within the economizer and the surrounding sensor faults.

\section{Results}\label{sec:results}

\begin{figure}
    \centering
    \includegraphics[width=0.9\linewidth]{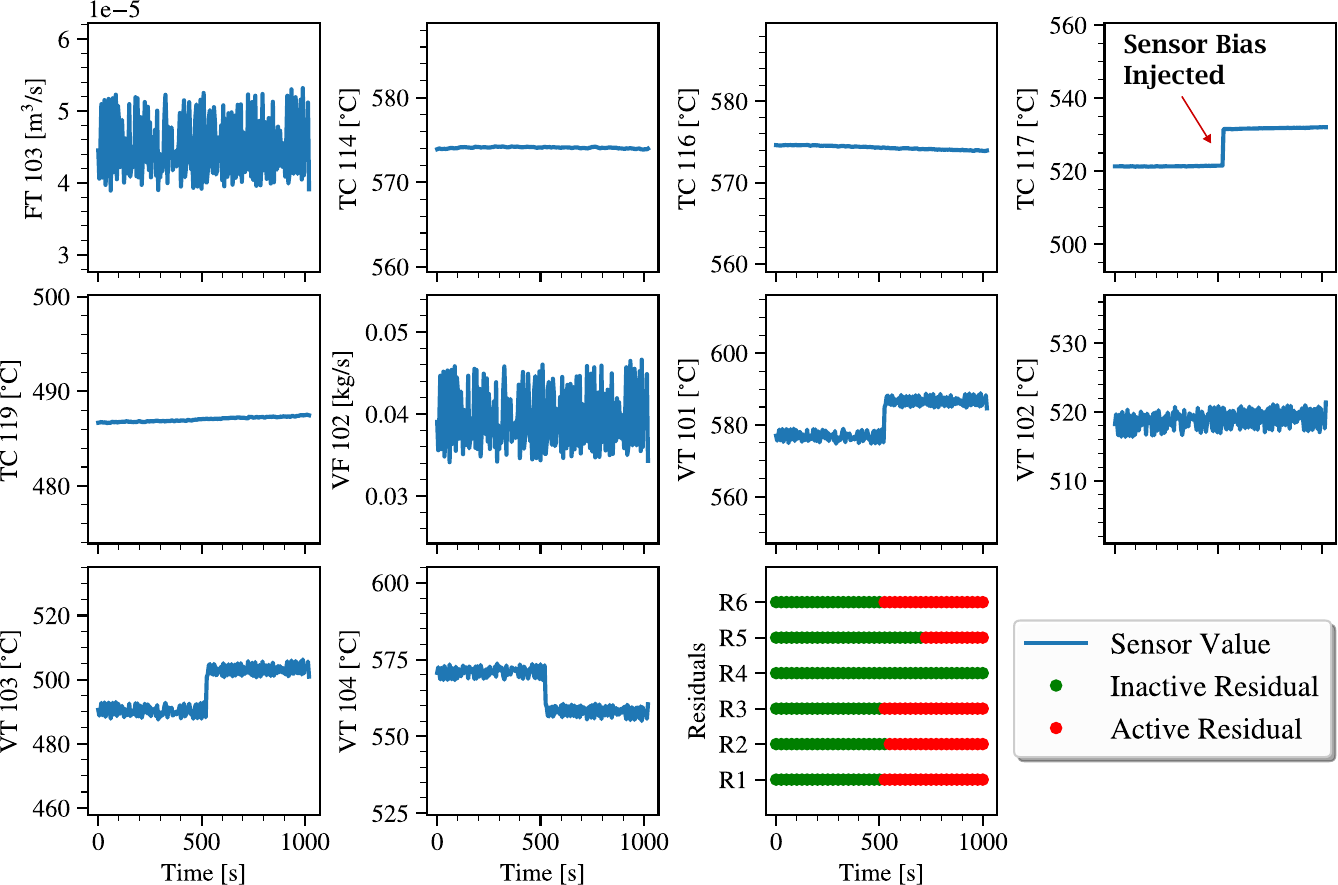}
    \caption{Sensor data and residual values during operation of the purification loop.
    The location of each sensor is presented in \cref{fig:METL}.
    An annotation in the TC 117 sensor subplot indicates the approximate time the facility operator injects a bias.}
    \label{fig:results}
\end{figure}

This section presents results from an experiment using the purification loop data.
The experiment involves injecting a fault in the thermocouple at the economizer's hot-side outlet (TC 117, shown in \cref{fig:METL}).
The fault is injected by adding a \SI{10.0}{\celsius} bias to the sensor output by operators of the METL facility. 
The collection of sensor data and PRO-AID residual evaluation is presented in \cref{fig:results}, where the fault is injected at approximately \SI{500}{\second}.
Several residuals (R1, R2, R3, R5, R6) are activated about a minute after injecting the bias.
The delay in residual activation is due to statistical significance requirements (\ie to exempt measurement noise from an inadvertent residual activation).
Using logical inference, the residual signature will correspond to a unique fault diagnosis.
In this case, PRO-AID would diagnose that a fault for sensor TC 117 has occurred based on the set of active residuals. 

Next, we detail the interactions with the Diagnostics Agent once the fault is introduced.
By calling the function \texttt{fault\_query}, the Agent employs logical reasoning to deduce the fault diagnosis and then shares this diagnosis with the user.
The agent's output is presented in \cref{lst:fault}.
Notably, the agent accurately identified fault F6, which is a sensor fault of TC 117.
Additionally, the agent indicates the unique set of sensors responsible for triggering the residuals listed.
This information aids operators in closely examining each implicated sensor.

\begin{lstlisting}[style=gptStyle, caption={Output of a fault query.}, label=lst:fault]
[agent.fault_query()]
[Output]
The fault signature identified is 'F6' which corresponds to the fault 'SensorFault-economizer.hot:temp:out'. 

The residuals identified by this fault are:
- r1: 'economizer-heat-balance_r' which relies on the sensors: ft_103, tc_114, tc_117, vf_102, tc_119, tc_116
- r2: 'economizer-heat-transfer_r' which relies on the sensors: ft_103, tc_114, tc_117, vf_102, tc_119, tc_116
- r3: 'economizer-heat-transfer_copy0_r' which relies on the sensors: ft_103, vt_101, tc_117, vf_102, tc_119, tc_116
- r5: 'economizer-heat-transfer_copy2_r' which relies on the sensors: ft_103, tc_114, tc_117, vf_102, vt_103, tc_116
- r6: 'economizer-heat-transfer_copy3_r' which relies on the sensors: ft_103, tc_114, tc_117, vf_102, tc_119, vt_104

The unique set of sensors involved in these residuals are: ft_103, tc_114, tc_116, tc_117, tc_119.
\end{lstlisting}

To aid explainability, the Operator must be able to query the Diagnostics Agent arbitrarily.
This capability is available with the \texttt{custom\_query} function.
In \cref{lst:custom1}, we ask the agent to explain the exoneration process.
The agent responds by explaining the exoneration process to the operator, in the context of fault F6 being activated.
There is an option to save the Agent's output in the context buffer.
This option would allow an Operator to continue their conversation with the Agent.

\begin{lstlisting}[style=gptStyle, caption={Output of a custom query.}, label=lst:custom1]
[agent.custom_query('Explain why the other faults were exonerated.')]
[Output]
The other faults were exonerated because they did not match the fault signature 'F6'. 

In this system, each fault has a unique signature that is identified by a specific set of residuals. When a fault occurs, it triggers certain residuals while leaving others inactive. The active residuals form the fault signature. 

In this case, the fault signature 'F6' corresponds to the fault 'SensorFault-economizer.hot:temp:out'. This fault triggered the residuals r1, r2, r3, r5, and r6. 

The other faults (F1, F2, F3, F4, F7, F8, F9) did not match this signature because they would have triggered a different set of residuals. Therefore, they were exonerated.
\end{lstlisting}

After being provided with a list of potentially faulty sensors, the Operator can benefit from examining the actual data these sensors recorded.
The function \texttt{query\_sensor\_data} facilitates this by allowing access to the relevant sensor data, with the analysis for TC 117 shown in \cref{lst:sensor}.
The Agent highlights anomalies in both statistical and spectral metrics.
To support this function, the \texttt{DependencyGraph} class maintains a buffer of past sensor readings.
An inbuilt function then analyzes various metrics from the batched data, with the results interpreted by the Agent.
\begin{lstlisting}[style=gptStyle, caption={Output of a sensor data query for TC 117.}, label=lst:sensor]
[agent.query_sensor_data('tc_117')]
[Output]
The statistical data for the sensor 'tc_117' show a sudden increase in both the mean and the standard deviation after the 11th data point. The rate of change in the mean and the standard deviation also show a significant increase at the same point. This could indicate an abnormal behavior or a possible fault in the system.

The spectral data show a decrease in the spectral entropy and an increase in the KL divergence after the 11th data point. This could indicate a change in the frequency distribution of the sensor data, which might be due to a fault or a change in the system's operation.

Overall, the metrics suggest that there might be a fault or an abnormal condition in the system that affects the 'tc_117' sensor after the 11th data point.
\end{lstlisting}
In contrast, the analysis of the readings from sensor TC 119, which was operating normally, is presented in \cref{lst:sensor_c}.
The Agent correctly determines that sensor TC 119 is operating correctly, based on the statistical and spectral metrics.
\begin{lstlisting}[style=gptStyle, caption={Output of a sensor data query for TC 119.}, label=lst:sensor_c]
[agent.query_sensor_data('tc_119')]
[Output]
The statistical data for the sensor 'tc_119' show a consistent increase in the mean value over time, with a slight decrease in the rate of change in the mean. The standard deviation also shows an increasing trend, indicating that the sensor readings are becoming less consistent over time.

The spectral data show a decreasing trend in the spectral entropy, which suggests that the frequency distribution of the sensor data is becoming less uniform. The KL divergence, is initially increasing but then starts to decrease after the 11th data point, indicating a change in the sensor data's predictability.

Overall, the metrics suggest that the sensor 'tc_119' is operating normally.
\end{lstlisting}

\section{Concluding Remarks}
Explainability in diagnostic tools for complex systems is crucial, enabling operators to comprehend not only the presence of a fault but also its origins and implications.
Purely data-driven approaches may fall short in providing useful explainability; physics-model-based diagnostic tools offer a more effective solution with their inherent causal relationship mapping.
Incorporating an LLM enhances this by translating the technical details from the physics-based model into understandable explanations for operators, and accommodating arbitrary queries about the system. 
However, care must be taken to constrain the LLM to prevent the dissemination of inaccurate information.

In this study, we introduced a system that creates an Agent to explain output from a physics-based diagnostic tool, PRO-AID.
Through data from a molten salt facility, we demonstrate that the Agent can explain the relationships between faults diagnosed and the sensors involved, respond to arbitrary queries from the operator, and analyze historical sensor measurement anomalies.

\section*{Acknowledgments}

This work was supported by the U.S. Department of Energy, Office of Nuclear Energy.
The authors would like to acknowledge Derek Kultgen for assistance in obtaining data from the METL facility.

\bibliographystyle{ans}
\bibliography{references.bib}

\begin{thebibliography}{1}
\newcommand{\enquote}[1]{``#1''}

\bibitem{choo2018visual}
\MakeUppercase{J.~Choo} and \MakeUppercase{S.~Liu}, \enquote{Visual analytics
  for explainable deep learning,} \emph{IEEE computer graphics and
  applications}, \textbf{38}, \emph{4}, 84--92 (2018).

\bibitem{hong2020remaining}
\MakeUppercase{C.~W. Hong}, \MakeUppercase{C.~Lee}, \MakeUppercase{K.~Lee},
  \MakeUppercase{M.-S. Ko}, \MakeUppercase{D.~E. Kim}, and
  \MakeUppercase{K.~Hur}, \enquote{Remaining useful life prognosis for turbofan
  engine using explainable deep neural networks with dimensionality reduction,}
  \emph{Sensors}, \textbf{20}, \emph{22}, 6626 (2020).

\bibitem{Langley_2017}
\MakeUppercase{P.~Langley}, \MakeUppercase{B.~Meadows},
  \MakeUppercase{M.~Sridharan}, and \MakeUppercase{D.~Choi},
  \enquote{Explainable Agency for Intelligent Autonomous Systems,}
  \emph{Proceedings of the AAAI Conference on Artificial Intelligence},
  \textbf{31}, \emph{2}, 4762--4763 (Feb. 2017).

\bibitem{NGUYEN2020107767}
\MakeUppercase{T.~N. Nguyen}, \MakeUppercase{T.~Downar}, and
  \MakeUppercase{R.~Vilim}, \enquote{A probabilistic model-based diagnostic
  framework for nuclear engineering systems,} \emph{Annals of Nuclear Energy},
  \textbf{149}, 107767 (2020).

\bibitem{DEKLEER200325}
\MakeUppercase{J.~De~Kleer} and \MakeUppercase{J.~Kurien},
  \enquote{Fundamentals of model-based diagnosis,} \emph{IFAC Proceedings
  Volumes}, \textbf{36}, \emph{5}, 25--36 (2003).

\bibitem{NGUYEN2022109002}
\MakeUppercase{T.~N. Nguyen}, \MakeUppercase{R.~Ponciroli},
  \MakeUppercase{P.~Bruck}, \MakeUppercase{T.~C. Esselman},
  \MakeUppercase{J.~A. Rigatti}, and \MakeUppercase{R.~B. Vilim}, \enquote{A
  digital twin approach to system-level fault detection and diagnosis for
  improved equipment health monitoring,} \emph{Annals of Nuclear Energy},
  \textbf{170}, 109002 (2022).

\bibitem{osti23035374}
\MakeUppercase{R.~Vilim}, \MakeUppercase{A.~Grelle}, \MakeUppercase{R.~Lew},
  \MakeUppercase{T.~Ulrich}, \MakeUppercase{R.~Boring}, and
  \MakeUppercase{K.~Thomas}, \enquote{Computerized operator support system and
  human performance in the control room,} in \enquote{Proceedings of the 10th
  International Topical Meeting on Nuclear Plant Instrumentation, Control, and
  Human-Machine Interface Technologies (NPIC \& HMIT 2017),}  (2017), pp.
  1195--1204.

\bibitem{osti_1492054}
\MakeUppercase{D.~Kultgen}, \MakeUppercase{C.~Grandy},
  \MakeUppercase{M.~Hvasta}, \MakeUppercase{D.~Lisowski},
  \MakeUppercase{W.~Toter}, and \MakeUppercase{A.~Borowski},
  \enquote{Mechanisms Engineering Test Loop-Phase 1 Status Report,} Tech. rep.,
  Argonne National Laboratory (ANL), Lemont, IL (United States) (2016).

\end{thebibliography}

\end{document}